\documentclass{article}
\usepackage[preprint]{corl_2026}
\usepackage{multicol}
\usepackage[dvipsnames,table]{xcolor}
\usepackage{graphicx}
\usepackage{caption}
\usepackage{wrapfig}
\usepackage{xspace}
\usepackage{amsmath,amssymb}
\usepackage{amsfonts}
\usepackage{scalefnt}
\usepackage{subcaption}
\usepackage{algorithmic}
\usepackage{algorithm}
\usepackage{booktabs}
\usepackage{etoolbox}

\definecolor{MondrianRed}{RGB}{176, 24, 16} 
\definecolor{DeepRed}{RGB}{184, 25, 17}
\definecolor{TsinghuaPurple}{RGB}{103, 19, 114}

\usepackage{cleveref}

\crefname{figure}{Fig.}{Figs.}
\Crefname{figure}{Fig.}{Figs.}
\crefname{table}{Table}{Tables}
\Crefname{table}{Table}{Tables}
\crefname{equation}{Eq.}{Eqs.}
\Crefname{equation}{Eq.}{Eqs.}

\hypersetup{
    colorlinks=true,
    linkcolor=TsinghuaPurple!70,
    citecolor=TsinghuaPurple!70,
    urlcolor=TsinghuaPurple!70,
    pdftitle={IMPACT: Learning Internal-Model Predictive Control for Forceful Robotic Manipulation},
    pdfauthor={Jiawei Gao, Chaoqi Liu, Peilin Wu, Haonan Chen, Yilun Du},
    pdfsubject={Preprint},
    pdfkeywords={Robot learning, Forceful manipulation, Predictive control}
}

\newcommand{\algofont}[1]{{\scalefont{1.1}{\texttt{\textbf{#1}}}}}
\newcommand{\method}{{\color{MondrianRed}\algofont{IMPACT}}\xspace}

\begin{document}

\title{IMPACT: Learning Internal-Model Predictive Control for Forceful Robotic Manipulation}

\author{
  \small Jiawei Gao$^{1}$ \quad Chaoqi Liu$^{1}$ \quad Peilin Wu$^{1}$ \quad Haonan Chen$^{1,2}$ \quad Yilun Du$^{1}$\\[0.4em]
  $^{1}$Harvard University \quad $^{2}$Stanford University\\[0.4em]
  \href{https://gao-jiawei.com/IMPACT/}{\textcolor{TsinghuaPurple}{\texttt{https://gao-jiawei.com/IMPACT/}}}
}
\date{}

\makeatletter
\apptocmd{\@maketitle}{%
  \par\bigskip
  \begin{center}
    \includegraphics[width=\textwidth]{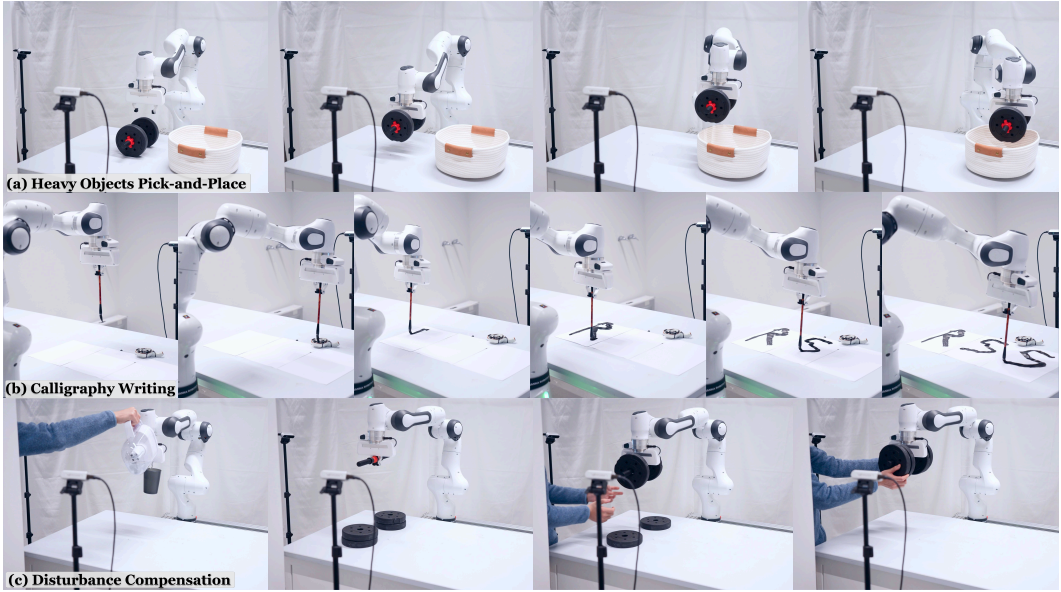}
    \captionof{figure}{We demonstrate the effectiveness of our framework across a variety of forceful manipulation tasks. (a) In heavy-object pick-and-place tasks (2.5\,kg and 5.0\,kg), the robot must generate forces that compensate for object gravity. (b) In calligraphy writing tasks, the robot is required to apply forces in the surface-normal direction of the table while simultaneously commanding motion in the tangential direction. (c) In disturbance compensation tasks, the robot must complete the task while adapting to varying object dynamics.}
    \label{fig:firstpage}
  \end{center}
}{}{\PackageWarning{titlepatch}{Failed to patch \string\@maketitle}}
\makeatother

\maketitle

\begin{abstract}
Real-world robotic manipulation tasks often involve forceful interactions with the environment, such as using tools of varying weights, transporting objects with different masses, and performing contact-rich tasks like table wiping.
Previous learning-based approaches typically employ imitation learning policies that output target end-effector poses tracked by low-level impedance controllers. In these systems, forceful interactions are either \textit{implicitly} realized through steady-state tracking errors or \textit{explicitly} commanded using wrist force/torque or tactile sensors.
However, implicit approaches generalize poorly across object weights, while explicit approaches require specialized hardware and increase system complexity.
In this work, we propose \method, a framework that decouples these forceful tasks into task-planning and internal-model-based predictive control.
Extensive simulation and real-world experiments demonstrate that the proposed framework achieves higher success rates and improved generalization to unseen object weights, as well as better safety and energy efficiency.

\end{abstract}


\section{Introduction}

\begin{wrapfigure}[15]{r}{0.5\textwidth}
  \centering
  \vspace{-1.5em}
  \includegraphics[width=\linewidth]{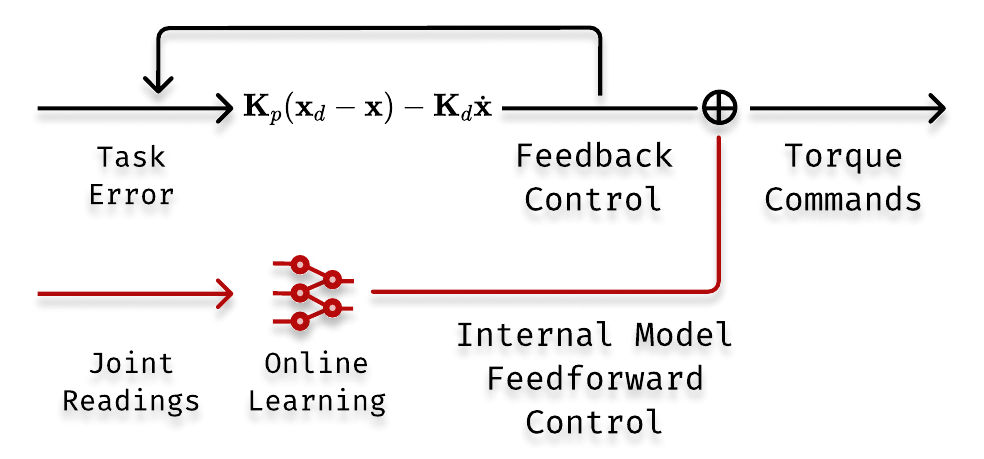}
  \caption{\textbf{Control diagram of our framework}. A task-space impedance controller provides feedback regulation based on tracking error, while an \textit{internal model} learns online from joint measurements to generate feedforward torque commands that compensate for persistent environment-induced interaction forces.}
  \label{fig:control_diagram}
\end{wrapfigure}

General-purpose robot learning systems hold the promise of enabling robots to acquire a wide range of manipulation skills, to autonomously learn from their experience and mistakes, and to adapt flexibly to new environments and scenarios.
Learning-based approaches address this goal by training visuomotor policies~\cite{chi2025diffusion, ze20243d, reuss2023goal, wolf2025diffusion, chen2025multi, zitkovich2023rt, kim2024openvla, høeg2025hybriddiffusion} using imitation learning algorithms~\cite{hussein2017imitation, chen2025tool, zare2024survey, hou2024diffusion, liu2025flexible} on human demonstration datasets~\cite{khazatsky2024droid, intelligence2025pi, o2024open}.
The policies are trained to imitate the demonstrated behaviors, learning to generate appropriate action sequences conditioned on observations.

Typically, ``actions" are represented as target end-effector poses or joint position commands, which are tracked by a low-level position controller~\cite{chi2025diffusion, chi2024universal}.
However, in scenarios involving forceful interactions with the environment, such as manipulating heavy objects, using tools~\cite{wang2024poco, chen2025tool}, or performing tasks that require significant contact forces, the policies need to not only specify desired kinematic trajectories, but also generate the necessary task-relevant interaction forces.
Previous methods address this issue by either \textbf{explicitly} commanding force-aware actions with information from force-torque sensors~\cite{liu2025factr, xue2025reactive, hou2025adaptive, chen2025multi}, or \textbf{implicitly} learning to modulate forces from datasets or reward functions~\cite{zhang2021learning, portela2024learning, zhi2025learning}.
However,  the \textit{explicit} approach requires accurate force--torque sensors mounted at the robot’s wrist, increasing system complexity.
For the \textit{implicit} approach, the policy must generate appropriate forces from the demonstration data, which couples task planning with object-dependent dynamics.
Take the task of lifting objects of varying weights as an example: the policy must learn to overshoot and create a steady-state end-effector tracking error to compensate for the object gravity.
This approach not only makes the learning problem more difficult, requiring substantially more data, but also leads to poor generalization to out-of-distribution object properties.

On the other hand, humans demonstrate robust and adaptive motor control capabilities when manipulating objects of varying weights and dynamics~\cite{todorov2004optimality}.
Humans do not have accurate force-torque sensors at their wrists,
yet they can effortlessly lift and manipulate objects with widely different weights and dynamics.
The key mechanism behind this ability is that the \textbf{cerebellum} builds \textbf{internal models} of the body and external objects based on prior experience~\cite{marr1969theory, wolpert1998internal, imamizu2012cerebellar}, and utilizes these models to generate appropriate feedforward forces to counteract predictable disturbances such as object gravity~\cite{franklin2008cns, franklin2007endpoint, bastian2006learning, pisotta2014cerebellar, burdet2001central}.

Inspired by this, we propose \method, an \textbf{I}nternal-\textbf{M}odel \textbf{P}redictive \textbf{C}on\textbf{T}rol framework for forceful robotic manipulation tasks.
The core idea is that, rather than explicitly measuring interaction forces with mounted force-torque sensors, we use joint torque readings and the manipulator equation to estimate interaction forces, and use online gradient descent to learn an internal model that predicts these interaction forces based on the history of states and actions. 
During task execution, the learned internal model is then used to compensate for the predicted interaction forces in a feedforward manner within the real-time control loop.
The illustration of the control diagram is shown in \cref{fig:control_diagram}.

We evaluate our framework on a suite of forceful manipulation tasks, including lifting objects of varying weights, compensating for external disturbances, and performing hybrid position--force control tasks such as calligraphy writing.
To systematically study the effectiveness of the proposed method compared to baseline methods, we performed controlled experiments in simulation, where we varied the object-dynamics properties during testing to evaluate the generalization ability of different methods.
We further validate our framework on a real-world Franka FR3 robot, demonstrating its capability to handle a wide range of forceful manipulation tasks in real-world settings.

\begin{figure*}[t]
  \centering
  \includegraphics[width=0.8\linewidth]{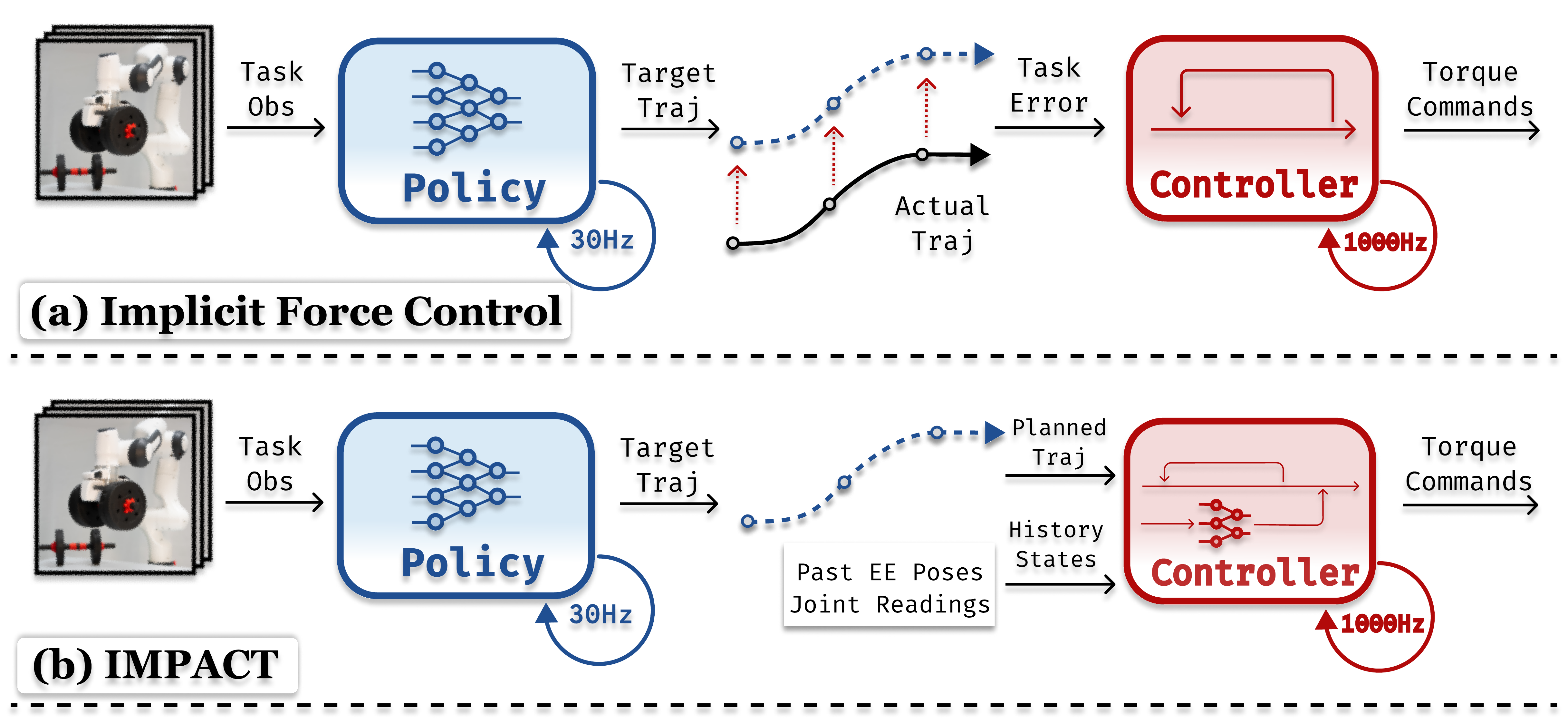}
  \caption{\textbf{Overview of \method and comparison with implicit force control method}. (a) In implicit force control, forces are generated implicitly by the policy through producing virtual target trajectories that induce tracking errors, which are converted into interaction forces by a low-level impedance controller. (b) In \method, the controller generates desired interaction forces: an internal model predicts contact and payload-induced forces and compensates via feedforward commands.}
  \vspace{-1.5em}
  \label{fig: pipeline}
\end{figure*}

\section{Methods}
\label{sec:methods}

We consider forceful robotic manipulation tasks under varying environment dynamics, including lifting objects with different weights and hybrid position--force tasks such as table wiping.
These tasks require the system to jointly generate motion trajectories and interaction forces.
As described in \cref{sec:method_background}, our framework uses a diffusion policy~\cite{chi2025diffusion} to imitate demonstration data and produce desired future end-effector trajectories, while interaction forces are handled online by the controller.

The core idea of our method is to \emph{learn an internal model that predicts and compensates for slowly varying external wrenches}, such as payload-induced forces, while maintaining compliant behavior during contact.
External wrenches are estimated from joint measurements (\cref{sec:wrench_estimation}) and decomposed into fast transients and slowly varying components using a first-order low-pass filter.
The resulting slow-varying wrench, which captures persistent effects such as object weight or steady contact forces, is used as the supervision signal for training the internal model.

During interaction, the internal model consumes a short history of joint states and end-effector measurements and is trained online via gradient descent to predict the slow-varying external wrench (\cref{sec:internal_model}).
This internal model is critical because the filtered wrench estimate converges slowly by design in order to suppress noise.
Once learned, the internal model enables the controller to immediately apply appropriate feedforward compensation when encountering a previously seen payload, without waiting for the slow wrench estimate to converge.
The pseudo-code of the overall control loop is shown in \cref{algo:policy_controller_core} in the Appendix.
\subsection{Background: Imitation Learning Policy Training and Impedance Control}
\label{sec:method_background}

We train an imitation-learning policy $\pi_\theta$ that outputs a horizon of desired end-effector target poses $\mathbf{X}^d_{t:t+H_p}$ conditioned on an observation history $\mathcal{O}_{t-H_o+1:t}$.
At inference time, target poses are executed in a receding-horizon fashion.
The planned trajectories are then sent to a low-level controller to generate joint torque commands $\boldsymbol{\tau}$.
Specifically, let $\mathbf{x}=(\mathbf{p},\mathbf{R})\in SE(3)$ denote the end-effector pose, with position $\mathbf{p}\in\mathbb{R}^3$ and orientation $\mathbf{R}\in SO(3)$, and let $\mathbf{x}_d$ denote the desired end-effector pose.
The task-space pose error is defined as $\mathbf{e}=\begin{bmatrix}\mathbf{p}_d - \mathbf{p} \\ \mathrm{Log}\!\left(\mathbf{R}_d \mathbf{R}^\top\right)\end{bmatrix}$, where $\mathrm{Log}(\cdot)$ maps a rotation matrix to its axis-angle representation.
A task-space impedance controller generates the desired interaction wrench
$\mathbf{w}_{\mathrm{imp}}=\mathbf{K}\mathbf{e}-\mathbf{D}\dot{\mathbf{x}}$,
where $\mathbf{K}$ and $\mathbf{D}$ are positive-definite stiffness and damping matrices, respectively, and $\dot{\mathbf{x}}$ denotes the end-effector spatial velocity.
The robot dynamics are governed by the manipulator equation $\mathbf{M}(\mathbf{q}) \ddot{\mathbf{q}} + \mathbf{C}(\mathbf{q},\dot{\mathbf{q}})\dot{\mathbf{q}} + \mathbf{g}(\mathbf{q}) = \boldsymbol{\tau} + \boldsymbol{\tau}_{\mathrm{ext}}$,
where $\mathbf{q}$, $\dot{\mathbf{q}}$, and $\ddot{\mathbf{q}}$ denote joint positions, velocities, and accelerations, respectively.
For tasks involving forceful interactions, the external wrench $\mathbf{w}_{\mathrm{ext}}$ at the end effector induces joint torques $\boldsymbol{\tau}_{\mathrm{ext}} = \mathbf{J}^\top \mathbf{w}_{\mathrm{ext}}$, where $\mathbf{J}$ is the end-effector Jacobian.

\subsection{Estimating External Wrenches with Joint Torque Residuals}
\label{sec:wrench_estimation}

Unmodeled interactions, including payload weight and contact forces, induce an external wrench
$\mathbf{w}_{\mathrm{ext}}\in\mathbb{R}^6$ at the end effector, which produces joint torques: $\boldsymbol{\tau}_{\mathrm{ext}} = \mathbf{J}(\mathbf{q})^\top \mathbf{w}_{\mathrm{ext}}$.
Using measured joint torques $\boldsymbol{\tau}_{\mathrm{meas}}$, we compute a model-based torque prediction $\boldsymbol{\tau}_{\mathrm{model}} = \mathbf{M}(\mathbf{q})\hat{\ddot{\mathbf{q}}} + \mathbf{C}(\mathbf{q},\dot{\mathbf{q}})\dot{\mathbf{q}} + \mathbf{g}(\mathbf{q}) + \boldsymbol{\tau}_{\mathrm{fric}}(\dot{\mathbf{q}})$, where $\hat{\ddot{\mathbf{q}}}$ is estimated from encoder measurements.
The torque residual $\boldsymbol{\tau}_\mathrm{res} = \boldsymbol{\tau}_{\mathrm{meas}} - \boldsymbol{\tau}_{\mathrm{model}}$ therefore captures the effect of unmodeled external interactions.
The external wrench estimate is obtained by solving a regularized least-squares problem:
\begin{equation}
\hat{\mathbf{w}}_{\mathrm{ext}}
=
\arg\min_{\mathbf{w}}
\left\|\mathbf{J}^\top\mathbf{w}-\boldsymbol{\tau}_\mathrm{res}\right\|_2^2
+ \lambda^2\|\mathbf{w}\|_2^2,
\end{equation}

yielding the closed-form solution $\hat{\mathbf{w}}_{\mathrm{ext}} = \left(\mathbf{J}\mathbf{J}^\top+\lambda^2\mathbf{I}\right)^{-1} \mathbf{J}\boldsymbol{\tau}_\mathrm{res}$.
However, the resulting estimate $\hat{\mathbf{w}}_{\mathrm{ext}}$ is inherently noisy due to modeling errors, sensor noise, and unmodeled high-frequency interactions.
To isolate the component suitable for prediction and feedforward compensation, we maintain a slowly varying estimate $\mathbf{w}_{\mathrm{slow}}$ using a first-order low-pass filtering update,
\begin{equation}
\mathbf{w}_{\mathrm{slow}}(t+\Delta t)
=
\mathbf{w}_{\mathrm{slow}}(t)
+
\eta\,\mathrm{clip}\!\left(
\hat{\mathbf{w}}_{\mathrm{ext}}(t)-\mathbf{w}_{\mathrm{slow}}(t)
\right),
\label{eq:slow_update}
\end{equation}
where $\eta\in(0,1)$ is the adaptation rate and $\mathrm{clip}(\cdot)$ applies per-axis saturation.
This tracks persistent external wrenches while attenuating high-frequency disturbances.
The resulting slow wrench estimate serves as the supervision signal for training the internal model.

\subsection{Internal Model Learning and Feedforward Compensation}
\label{sec:internal_model}

The slow-varying external wrench $\mathbf{w}_{\mathrm{slow}}$ captures persistent interaction effects such as payload-induced forces or steady contact forces.
Rather than directly using $\mathbf{w}_{\mathrm{slow}}$ for compensation, which converges slowly by design, we introduce an internal model that learns to \emph{predict} this quantity from the robot’s recent state history.
Specifically, the internal model parameterizes a predictor $\mathbf{w}_{\mathrm{pred}}=\mathcal{M}_\phi\!\left(\mathbf{z}_{t-L:t}\right)$, 
where $\mathbf{z}_{t-L:t}$ denotes a short horizon of joint states and end-effector measurements, and $\mathcal{M}_\phi$ is trained online to approximate the slow-varying external wrench.
To improve robustness and prevent adaptation to noisy transients, the internal model is updated only when its prediction deviates significantly from the measured interaction forces.
Once the prediction error falls below the threshold, learning is suspended, and the internal model output is treated as a reliable prediction.

After learning, the internal model enables \textbf{predictive force compensation}.
When the robot encounters a previously learned interaction condition, the predicted wrench $\mathbf{w}_{\mathrm{pred}}$ is immediately applied as a feedforward command, $\mathbf{w}_{\mathrm{ff}} = \mathbf{w}_{\mathrm{pred}}$, without waiting for the slow external wrench estimate to converge.
This anticipatory compensation reduces steady-state tracking error and prevents large deflections in the impedance controller.
The feedforward wrench is then mapped to joint torques via
$\boldsymbol{\tau}_{\mathrm{ff}} = \mathbf{J}(\mathbf{q})^\top \mathbf{w}_{\mathrm{ff}}$,
and is added to the impedance control law as in \cref{algo:policy_controller_core}.
Substituting the total torque command into the robot dynamics, the closed-loop system obeys
\begin{equation}
\mathbf{M}(\mathbf{q}) \ddot{\mathbf{q}}
+
\mathbf{C}(\mathbf{q},\dot{\mathbf{q}})\dot{\mathbf{q}}
+
\mathbf{g}(\mathbf{q})
=
\boldsymbol{\tau}_{\mathrm{imp}}
+
\boldsymbol{\tau}_{\mathrm{ff}}
+
\boldsymbol{\tau}_{\mathrm{null}}
+
\boldsymbol{\tau}_{\mathrm{ext}},
\label{eq:closed_loop_dynamics}
\end{equation}
where $\boldsymbol{\tau}_{\mathrm{ext}} = \mathbf{J}(\mathbf{q})^\top \mathbf{w}_{\mathrm{ext}}$ represents joint torques induced by external interaction forces, $\boldsymbol{\tau}_{\mathrm{imp}}$ denotes the impedance feedback term, $\boldsymbol{\tau}_{\mathrm{null}}$ enforces secondary objectives, and $\boldsymbol{\tau}_{\mathrm{ff}}$ is the predictive feedforward compensation term.

\section{Experiments}

\begin{figure*}[t]
  \centering
  \includegraphics[width=0.9\linewidth]{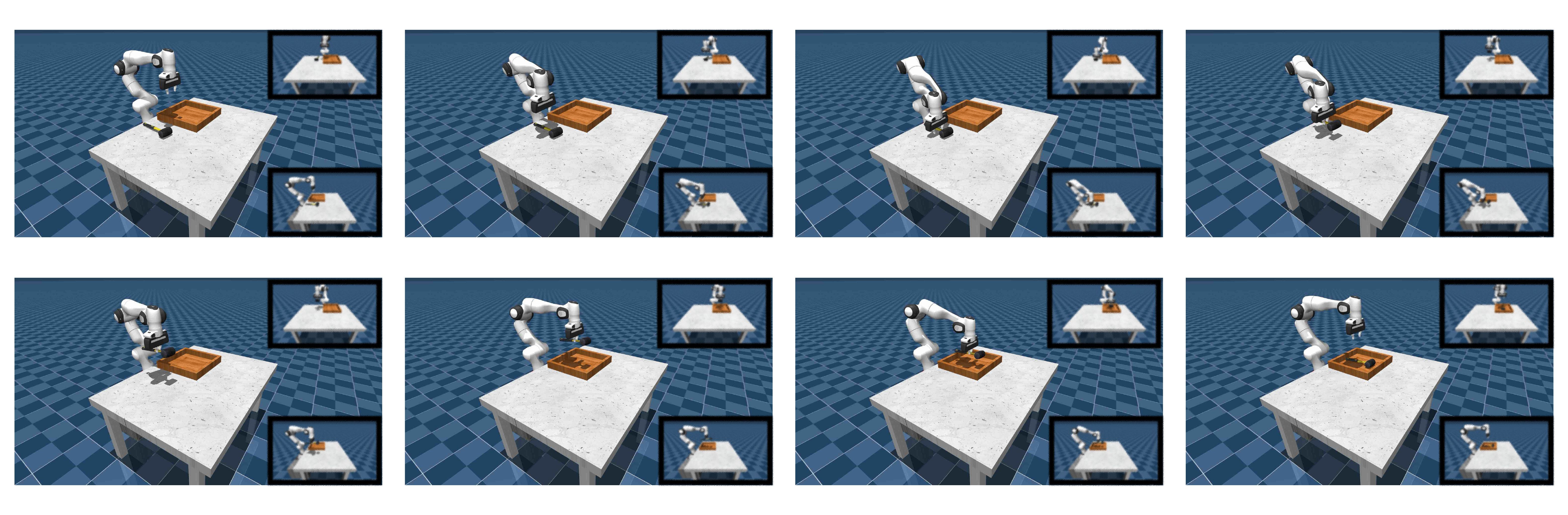}
  \caption{\textbf{Simulation Experiments Setup}. We set up tasks in the MuJoCo simulator for controlled evaluation of the proposed framework against baseline methods. The teleoperation, data-postprocessing, and policy training pipelines are consistent with the real-robot experiments, ensuring a fair comparison.}
  \label{fig:mujoco_setup}
\end{figure*}

To evaluate the effectiveness of \method for forceful robotic manipulation tasks, we conduct extensive experiments in both simulation and real-world settings.
Experiment results demonstrate that our framework achieves higher success rates and better out-of-distribution generalization capabilities compared to standard learning-based baselines, particularly in scenarios involving significant changes in object mass and dynamics.

\subsection{Experiments Setup}

\begin{wrapfigure}[17]{r}{0.5\textwidth}
  \vspace{-0.75em}
  \centering
  \includegraphics[width=0.8\linewidth]{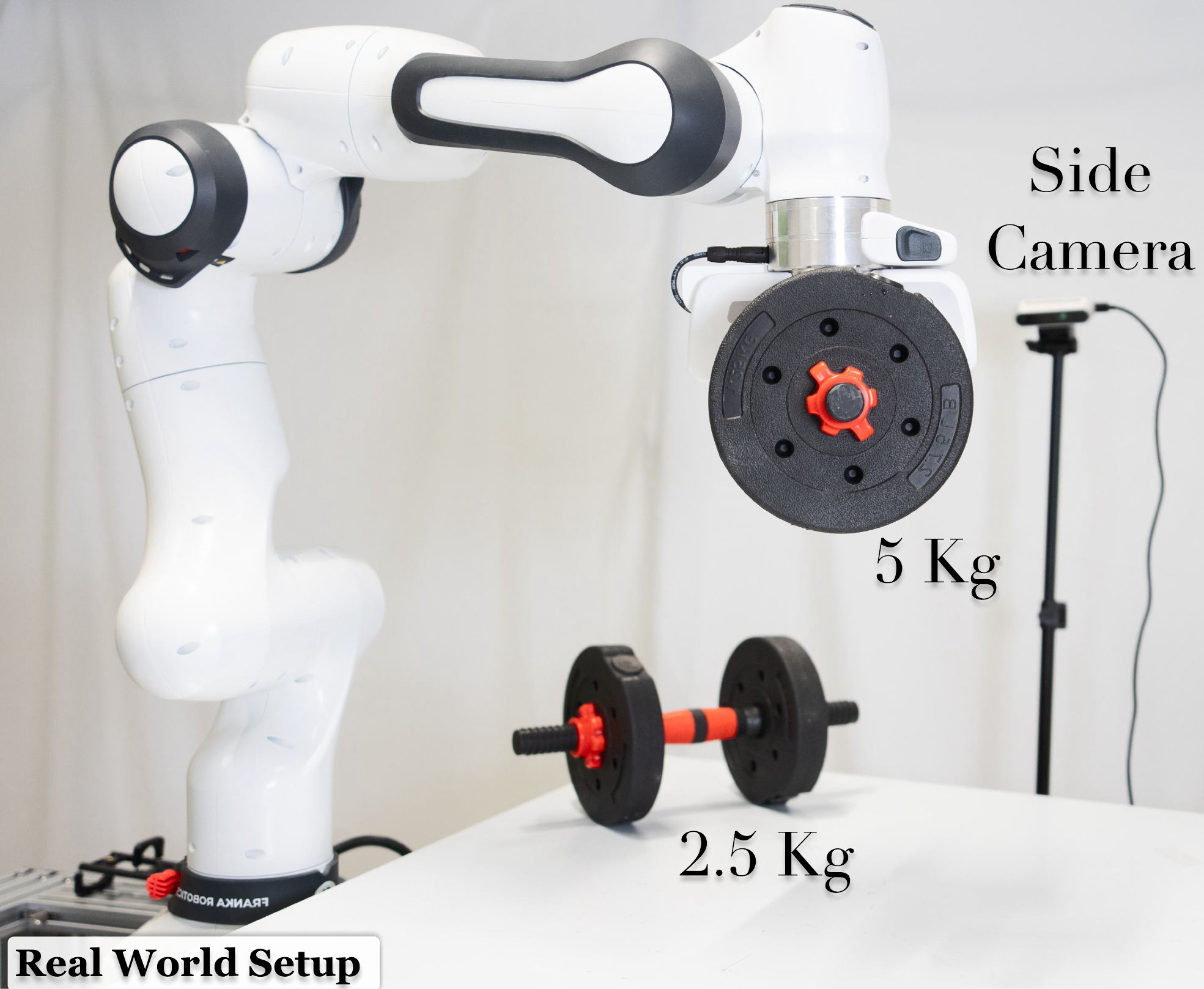}
  \caption{\textbf{Real-World Setup}. We use a Nintendo Switch Joy-Con to teleoperate the Franka FR3 robot to pick up dumbbells weighing 2.5\,kg and 5\,kg. The teleoperation data are collected at 30~Hz and used to train the policies.}
  \label{fig:real_world}
  \vspace{-0.75em}
\end{wrapfigure}

\paragraph{Real-World Experiments.}
We conduct real-world experiments using a 7-DOF Franka FR3 manipulator operating in a tabletop workspace monitored by two fixed-base RGB-D cameras (Fig.~\ref{fig:real_world}).
The action space is 8-dimensional, consisting of end-effector Cartesian pose commands and gripper actuation.
Observations and actions are synchronized and recorded at 30~Hz for policy training.
To evaluate the effectiveness of our framework under varying object dynamics, we conduct experiments using dumbbells weighing 2.5\,kg and 5\,kg.
Demonstration data are collected for a pick-and-place task using the 2.5\,kg dumbbell, and the trained framework is evaluated on both the 2.5\,kg and 5\,kg dumbbells without providing explicit information about the object weight.

\paragraph{Simulation Benchmark.}
We use the MuJoCo physics engine~\cite{mujoco} to simulate the Franka FR3 robot in a tabletop manipulation environment with two virtual RGB-D cameras (Fig.~\ref{fig:mujoco_setup}).
The scene contains task-relevant objects, including a hammer and a target basket, whose positions are randomized across episodes.
Expert demonstrations are generated via motion planning.
We use simulation to conduct controlled experiments with varying hammer mass, where the hammer mass is randomized from 0.1 to 10~kg, and evaluate the effectiveness of our framework through repeated multi-episode evaluations to ensure statistical significance.

\paragraph{Baseline Methods.}
We compare \method with the following learning-based baselines: 
1) \texttt{Vanilla DP.}
A standard Diffusion Policy (DP)~\cite{chi2025diffusion} trained using demonstrations collected at a single nominal object mass.
This corresponds to imitation learning policies trained only with light objects.
In MuJoCo, we fix the hammer mass to 0.2~kg and collect 200 demonstration episodes.
2)
\texttt{Augmented DP.}
To evaluate whether scaling demonstration data alone enables the policy to master the task, we collect demonstrations across a broad range of object masses using a domain-randomization strategy.
In MuJoCo, we randomly sample the hammer mass from 0.1 to 8~kg and collect 200 demonstration episodes.
In baseline methods, learned policies output target end-effector poses, which are tracked by the low-level impedance controller. The stiffness and damping parameters are fixed.

\subsection{Effectiveness of \texorpdfstring{\method}{IMPACT} in Simulation and Real-world Benchmarks}

\begin{figure*}[t]
  \centering
  \includegraphics[width=0.8\linewidth]{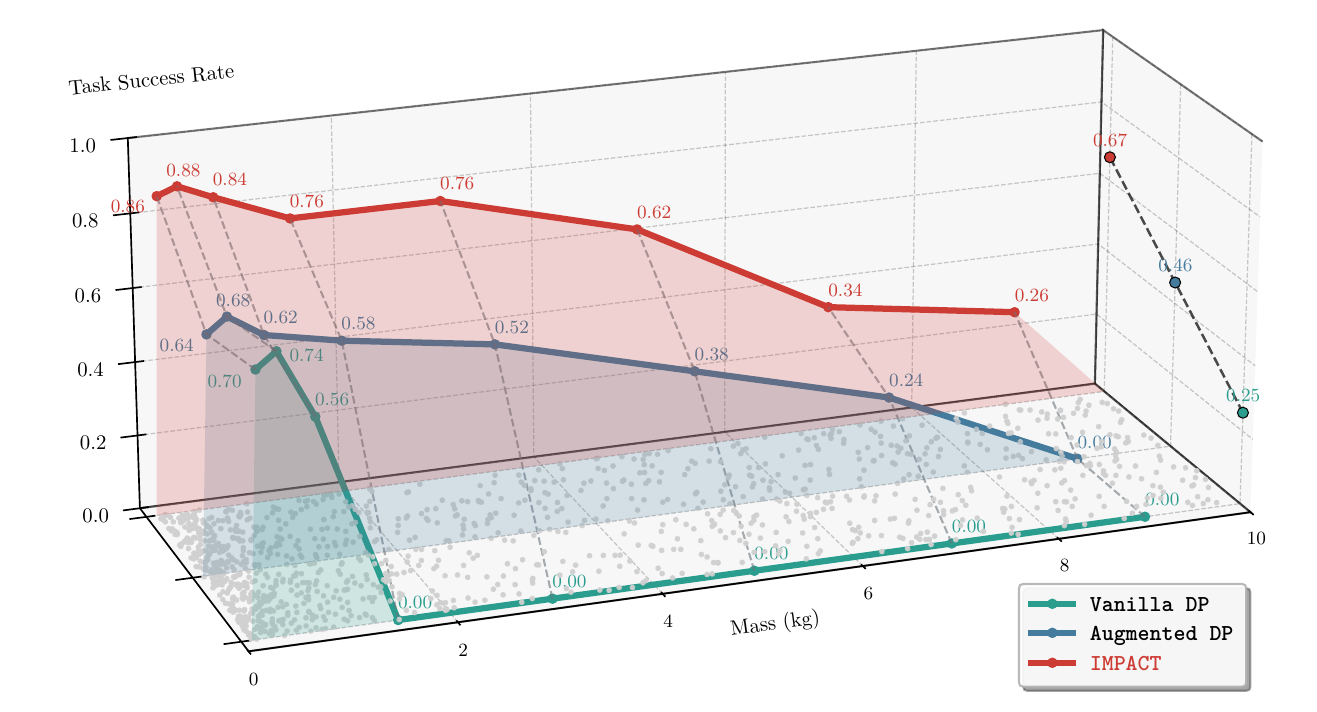}
  \caption{\textbf{Simulation Benchmarking.} Evaluation of task success rates in the MuJoCo simulation benchmark across varying object masses (0--10~kg). $x$ axis denotes the object mass, $z$ axis denotes the task success rate, and $y$ axis represents different methods. While \texttt{Vanilla DP} (trained on 0.2~kg) fails to generalize and \texttt{Augmented DP} (trained on 0.1--8.0~kg) degrades at high payload ($>$ 8~kg), \method maintains superior performance across the entire physical range despite being trained exclusively on the nominal mass.}
  \label{fig:sim_benchmarking}
\end{figure*}

\paragraph{Controlled experiments in Simulation.}
To rigorously evaluate the effectiveness of our approach, we conduct controlled experiments in the MuJoCo simulator, where we can systematically vary the object mass and evaluate the robustness of different methods to out-of-distribution dynamics.
Success rates across benchmarks are reported in Fig.~\ref{fig:sim_benchmarking}. 
In this setup, both \method and \texttt{Vanilla DP} are trained exclusively at a nominal object mass of 0.2~kg, whereas \texttt{Augmented DP} is trained from demonstrations collected across a wide range of masses (0.1--8.0~kg).

Our results show that the performance of \texttt{Vanilla DP} decreases as the object mass increases, indicating overfitting to the nominal training dynamics and limited generalization to changes in object mass.
\texttt{Augmented DP} improves robustness relative to the vanilla baseline by training on a wider range of masses, but its success rate still drops at higher masses, especially when the mass is outside the training distribution (e.g., $>8$~kg).
In contrast, \method maintains high success rates across the entire evaluated mass range and outperforms both baselines.
Notably, \method achieves better performance than \texttt{Augmented DP} despite having no prior exposure to varied object masses during training, indicating that explicitly modeling and compensating for payload dynamics leads to better generalization than relying solely on increased data diversity.

\paragraph{Real-world experiments on Franka-FR3 robot.}

We also evaluate our method on real-world forceful manipulation tasks using the Franka FR3 robot, as shown in Fig.~\ref{fig:real_world}.

\begin{wraptable}[8]{r}{0.5\textwidth}
  \vspace{-0.75em}
  \centering
  \setlength{\tabcolsep}{3.0pt}
  \resizebox{\linewidth}{!}{%
  \begin{tabular}{l|ccc}
  \toprule
  \textbf{Method} & \textbf{Train Mass (kg)} & \textbf{Test 2.5~kg} & \textbf{Test 5~kg} \\
  \midrule
  \texttt{Vanilla DP} & 2.5 & 18 / 25 & \phantom{0}0 / 25 \\
  \texttt{Augmented DP} & 2.5 + 5.0 & \phantom{0}9 / 25 & \phantom{0}5 / 25 \\
  \method & 2.5 & \textbf{22 / 25} & \textbf{21 / 25} \\
  \bottomrule
  \end{tabular}%
  }
  \caption{\textbf{Real-World Benchmarking.}
  Success rates when evaluating policies on 2.5\,kg and 5\,kg dumbbells under different training distributions.}
  \label{tab:payload_generalization}
  \vspace{-0.75em}
\end{wraptable}

The task involves picking up and placing dumbbells of varying mass (2.5~kg and 5~kg) on a table.
We compare the performance of \method against the two baselines, which are trained using demonstration data collected at the nominal mass (2.5~kg) and across a range of masses (2.5~kg and 5~kg), respectively.
Experiments shown in \Cref{tab:payload_generalization} indicate that \method outperforms both baselines in terms of success rates on both the nominal and heavier dumbbells, demonstrating generalization capabilities.

\subsection{Analysis and ablations of \method}
\label{sec:protocol}

\begin{wrapfigure}[25]{r}{0.5\textwidth}
  \vspace{-0.75em}
  \centering
  \includegraphics[width=\linewidth]{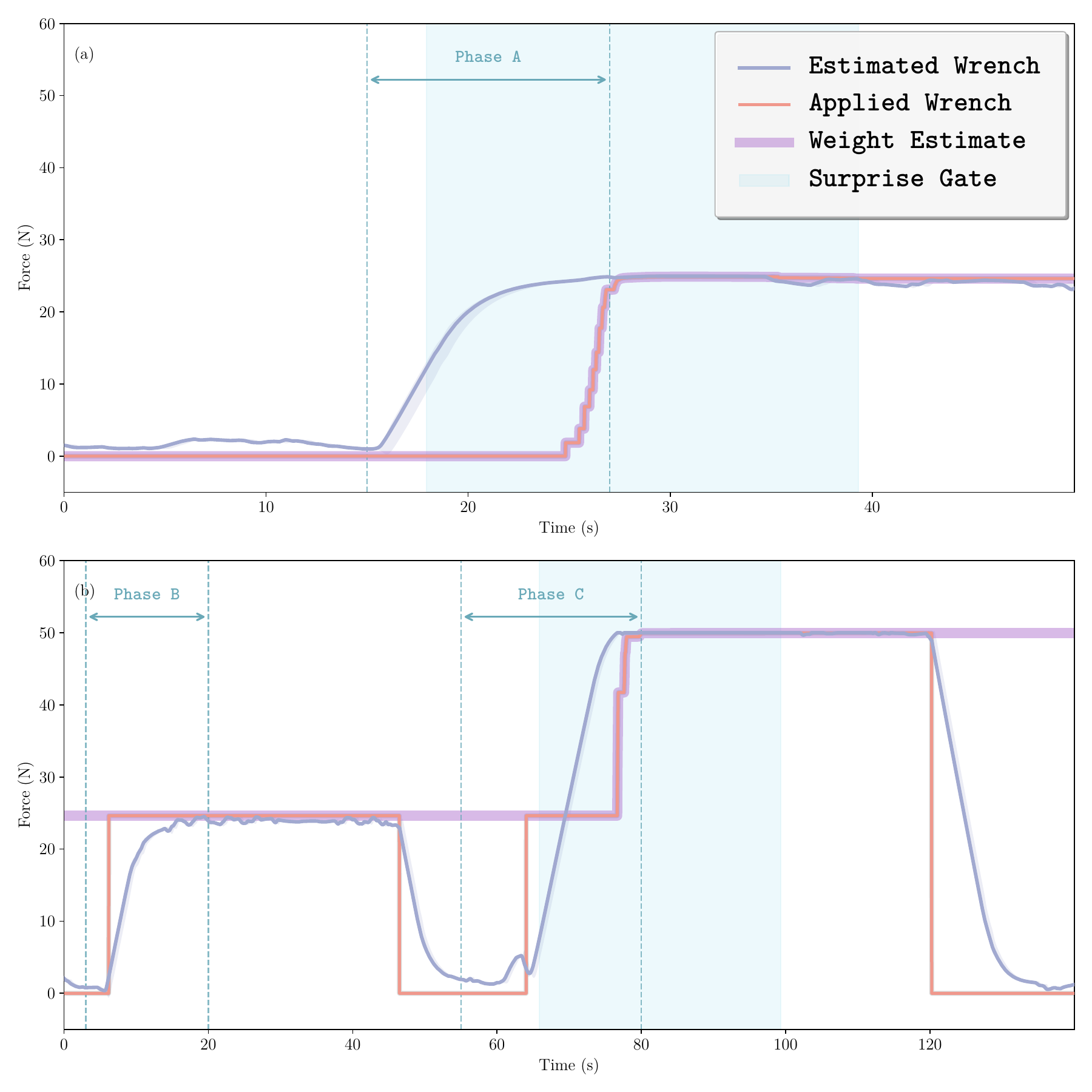}
  \caption{\textbf{Visualization of the key metrics during the task protocol}. We visualize the internal model behavior across three phases: (A) initial weight learning via surprise gate activation, (B) zero-delay feedforward compensation for a known load, and (C) gate reactivation for unexpected mass increase (2.5~kg to 5~kg).}
  \label{fig:force_evolution}
  \vspace{-0.75em}
\end{wrapfigure}

To understand the underlying mechanisms that contribute to the superior performance of \method, we conduct a detailed analysis of the internal model's behavior and its impact on control performance.
We design a task protocol that we use the policy trained on datasets collected at 2.5~kg to pick up the 2.5~kg dumbbell, let the controller learn and adapt online.
When the policy successfully finishes the pick-and-place task, we reset the environment, and let the policy finish this task again.
Then, we switch to the 5~kg dumbbell without any prior information  to the policy, and test the policy's ability to re-adapt to the new dynamics.
We record the key metrics, including the applied wrench, estimated external wrench, internal model predictions, and pose tracking error throughout the episode to analyze how \method adapts to changes in payload dynamics.
The overall visualization can be found in Fig.~\ref{fig:wrench_traj_vis}, and we provide zoom-in visualizations of the internal model behavior in Fig.~\ref{fig:force_evolution} and the pose tracking performance in Fig.~\ref{fig:pose_err_and_applied_wrench}.
We also test the performance of baseline methods on this task protocol using impedance control without feedforward compensation, and the results are shown in Fig.~\ref{fig:impedance_pose_err}.

When the policy first picks up the 2.5~kg dumbbell, the surprise gate activates due to the unexpected load, allowing the internal model to rapidly converge to an accurate estimate of the external wrench.
This corresponds to the \texttt{Phase A} in Fig.~\ref{fig:force_evolution}, where the internal model quickly learns to predict the external load, and the feedforward compensation effectively cancels the load, resulting in low pose error (Fig.~\ref{fig:pose_err_and_applied_wrench}).
This is essentially an online-learning process, where the internal model learns to predict the external load based on the observed discrepancies between the measured and expected torques, and the feedforward compensation allows the controller to maintain accurate tracking despite the presence of the load.
In the second stage, when the policy continues to perform the pick-and-place task with the 2.5~kg dumbbell, since the internal model has already built an accurate representation of the environmental dynamics, the surprise gate remains inactive, and the feedforward compensation continues to effectively cancel the external load, maintaining low pose error, which corresponds to \texttt{Phase B} in Fig.~\ref{fig:force_evolution}.
In the third stage, when the dumbbell is switched to 5~kg without any prior information, the internal model's internal dynamics representation still remains the case of 2.5~kg, which leads to a significant increase in the prediction error of the external load.
Thus, the surprise gate reactivates, allowing the internal model to rapidly update its predictions to account for the new dynamics, which corresponds to \texttt{Phase C} in Fig.~\ref{fig:force_evolution}.

This experiment demonstrates the key advantages of \method: 1) the ability to rapidly learn and adapt to new dynamics through the surprise gate mechanism, and 2) the ability to maintain accurate control performance through feedforward compensation based on internal model predictions, even in the presence of significant changes in object mass and dynamics.
In contrast, the baseline impedance controller without feedforward compensation (Fig.~\ref{fig:impedance_pose_err}) exhibits significant steady-state error across all phases due to the lack of load compensation, highlighting the importance of the internal model and feedforward compensation in achieving robust performance under varying dynamics.

\section{Related Work}

\paragraph{Position-Force Control for Robotic Manipulation}
Impedance control and hybrid position–force control~\cite{raibert1981hybrid, hogan1984impedance, mason2007compliance, yoshikawa2003dynamic, siciliano1999robot} are widely used for forceful robotic manipulation tasks.
In task and motion planning, desired interaction forces can be planned by creating virtual targets, with an impedance controller generating the corresponding forces~\cite{yashinski2024performing, holladay2024robust, holladay2021planning, holladay2019force, pasricha2025dynamics, guo2024flying}.
Recently, the community has begun to use learning-based visuomotor policies for forceful manipulation tasks.
Desired forces can be \textit{explicitly} commanded using force–torque sensors mounted at the robot end effector~\cite{hou2025adaptive, liu2025factr, xue2025reactive, xu2025compliant, geiger2025diffusion, lee2025manipforce, choi2026wild}: during data collection, interaction forces are recorded, and policies are trained to predict both target trajectories and the required interaction forces.
Alternatively, desired forces can be generated \textit{implicitly} by having the policy output virtual target trajectories, with a low-level impedance controller producing interaction forces through feedback control~\cite{zhang2021learning, ulmer2021learning, martin2019variable, abu2018force}.
However, the explicit methods require additional force sensors, increasing system cost and complexity, while implicit methods struggle to generalize to object dynamics outside the collected dataset~\cite{aljalbout2024shortcomings}.
Our method, in contrast, decouples high-level learning from low-level control: the high-level policy outputs planned trajectories, while the low-level controller uses an internal model to predict interaction forces and compensate for them via feedforward commands.

\paragraph{Motor Learning in Human Cerebellum and Internal Model Control}

Our work has connections to neuroscience hypotheses on motor control and the mechanisms of cerebellum.
The cerebellum is recognized as a key component of motor learning, enabling humans and animals to acquire skilled movements and adapt to novel dynamics~\cite{marr1969theory, kandel2000principles, albus1971theory, eccles2013cerebellum}.
The key mechanism behind this is, the cerebellum constructs an internal model of environmental dynamics and refines it through practice and adaptation~\cite{wolpert1998internal, imamizu2012cerebellar}.
This internal model supports predictive control~\cite{bastian2006learning, pisotta2014cerebellar}, allowing the nervous system to anticipate the sensory consequences of actions and produce smooth, accurate movements~\cite{burdet2001central, franklin2007endpoint, franklin2008cns}.
Motivated by these insights, cerebellum-inspired architectures have been explored across a range of robotic domains, including motion planning~\cite{tolu2020cerebellum, zahra2020vision}, locomotion~\cite{long2023hybrid, long2025learning}, multi-robot coordination~\cite{gao2024coohoi}, and skill learning for Vision–Language–Action models~\cite{tan2025roboos}.
In parallel, internal model control (IMC) in control theory~\cite{garcia1982internal} formalizes similar ideas and has been successfully applied to robotic manipulation tasks such as grip force modulation~\cite{kawato1999internal}, stiffness regulation~\cite{ganesh2010biomimetic}, and feedforward compensation in unstable interactions~\cite{yang2011human}.
Building on these principles, our approach focuses on leveraging cerebellar-style predictive compensation for robotic manipulation under varying and unknown object dynamics.
By learning an internal model of interaction forces, the controller generates feedforward force commands that adapt to changes in object properties, improving both control performance and learning efficiency.

\section{Conclusion} 
\label{sec:conclusion}

In this work, we presented \method, a framework for forceful robotic manipulation that decouples high-level visuomotor planning from low-level force compensation through an internal-model-based controller.
We evaluated the proposed framework in both simulation and real-world experiments across a range of forceful manipulation tasks, including lifting objects of varying weights, disturbance compensation, and hybrid position--force control tasks.
The results demonstrate that our approach significantly improves robustness and generalization to unseen object dynamics compared to implicit force control methods and data augmentation baselines, while maintaining accurate trajectory tracking and stable contact behavior.

\paragraph{Limitations}
However, our current framework can only build internal models of dynamic properties that are directly related to forceful interactions, such as object gravity, contact stiffness, and friction.
Our method cannot be directly applied to non-rigid-body objects, such as deformable objects, fluids, and granular materials, which require modeling of more complex dynamic properties, such as deformation, flow, and particle interactions.
To make this internal-model framework more general, future work will explore extending the internal model to handle more complex contact conditions and integrating richer sensory modalities for broader classes of manipulation tasks.


\bibliography{references}

\clearpage
\appendix
\section{Appendix}

We provide a detailed description of the \method framework, including the algorithmic implementation, hyper-parameters, and visualizations of the internal model learning and predictive control process in real-world experiments.
\subsection{Internal Model Learning and Predictive Control of \method}
\label{app:policy_controller}

We summarize the whole pipeline of \method in \cref{algo:policy_controller_core}, which includes the policy inference, impedance control, external wrench estimation, internal model prediction, and predictive compensation steps.
Note that this is for policy inference, the policy is trained following the standard diffusion policy training pipeline~\cite{chi2025diffusion}.

\begin{algorithm}[H]
\caption{\method framework}
\label{algo:policy_controller_core}
\begin{algorithmic}[1]
\REQUIRE Policy $\pi_\theta$; impedance gains $(\mathbf{K},\mathbf{D})$; observation history $\mathcal{O}_t$

\FOR{each policy update time $t_k$ (30Hz)}
  \STATE $\mathbf{X}^d_{t_k:t_k+H_p} \gets \pi_\theta(\mathcal{O}_{t_k})$

  \FOR{each control step $t \in [t_k,t_{k+1})$ (1000Hz)}
    \STATE $\mathbf{x}_d \gets \mathcal{S}(\mathbf{X}^d,t)$
    \STATE Compute pose error $\mathbf{e}$ and Jacobian $\mathbf{J}$
    \STATE $\mathbf{w}_{\mathrm{imp}} \gets \mathbf{K}\mathbf{e}-\mathbf{D}\mathbf{J}\dot{\mathbf{q}}$
    \STATE $\boldsymbol{\tau}_{\mathrm{imp}} \gets \mathbf{J}^\top \mathbf{w}_{\mathrm{imp}}$

    \STATE \textbf{(External wrench estimation)}
    \STATE $\mathbf{r} \gets \boldsymbol{\tau}_{\mathrm{meas}}-\boldsymbol{\tau}_{\mathrm{model}}$
    \STATE $\hat{\mathbf{w}}_{\mathrm{ext}} \gets (\mathbf{J}\mathbf{J}^\top+\lambda^2\mathbf{I})^{-1}\mathbf{J}\mathbf{r}$
    \STATE $\mathbf{w}_{\mathrm{slow}} \gets \mathrm{LowPass}(\hat{\mathbf{w}}_{\mathrm{ext}})$

    \STATE \textbf{(Internal model prediction)}
    \STATE $\mathbf{w}_{\mathrm{pred}} \gets \mathcal{M}_\phi(\mathbf{z}_{t-L:t})$

    \STATE \textbf{(Predictive compensation)}
    \STATE $\mathbf{w}_{\mathrm{ff}} \gets \mathbf{w}_{\mathrm{pred}}$
    \STATE $\boldsymbol{\tau}_{\mathrm{ff}} \gets \mathbf{J}^\top \mathbf{w}_{\mathrm{ff}}$

    \STATE \textbf{(Torque command)}
    \STATE $\boldsymbol{\tau} \gets \boldsymbol{\tau}_{\mathrm{imp}}+\boldsymbol{\tau}_{\mathrm{ff}}+\boldsymbol{\tau}_{\mathrm{null}}$
    \STATE Apply $\boldsymbol{\tau}$
  \ENDFOR
\ENDFOR
\end{algorithmic}
\end{algorithm}

The key hyperparameters used by the policy, controller, wrench estimator, and internal model are summarized in \cref{tab:hyperparameters}.

\begin{table}[h]
\centering
\small
\setlength{\tabcolsep}{4pt}
\begin{tabular}{p{0.30\linewidth}p{0.13\linewidth}p{0.48\linewidth}}
\toprule
\textbf{Hyperparameter} & \textbf{Symbol} & \textbf{Value} \\
\midrule
Policy prediction horizon & $H_p$ & 32 \\
Observation history length & $H_o$ & 2 \\
Impedance stiffness gains & $\mathbf{K}$ & $\mathrm{diag}(400,400,400,\allowbreak 30,30,30)$ \\
Impedance damping gains & $\mathbf{D}$ & $\mathrm{diag}(40,40,40,\allowbreak 10.95,10.95,10.95)$ \\
Wrench regularization weight & $\lambda$ & 0.02 \\
Low-pass adaptation rate & $\eta$ & force: 2.0 s$^{-1}$, torque: 1.0 s$^{-1}$ \\
Low-pass filter time constant & -- & 0.0495 s \\
Clip saturation bounds & -- & 10 N, 1 Nm\\
Internal model history length & $L$ & 32 \\
Internal model learning rate & -- & 0.01 \\
Internal model update threshold & -- & surprise $\geq$ 5 N; slow-wrench rate $\leq$ 0.2\\
Policy update frequency & -- & 30 Hz\\
Controller frequency & -- & 1000 Hz \\
\bottomrule
\end{tabular}
\caption{Hyperparameters used by the policy, controller, wrench estimator, and internal model.}
\label{tab:hyperparameters}
\end{table}

\subsection{Additional Real-World Visualizations}
\label{app:real_world_visualizations}

\begin{figure}[h]
  \centering
  \includegraphics[width=0.8\linewidth]{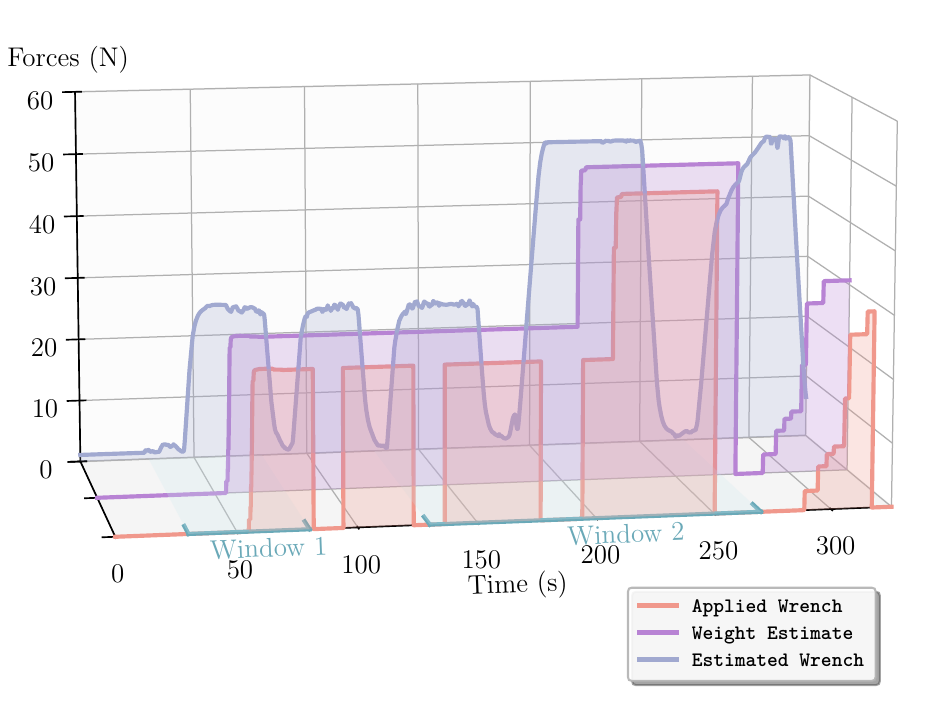}
  \caption{\textbf{Example visualization of one real-world episode.} 3D visualization of applied wrench, estimated wrench, and internal weight estimates. \method demonstrates successful mass identification in Window 1 (2.5~kg) and rapid re-adaptation in Window 2 (5~kg) upon detecting load discrepancies.}
  \label{fig:wrench_traj_vis}
\end{figure}

We record the key metrics during real-world experiments, including the applied wrench, estimated external wrench, and internal model predictions, as described in \cref{sec:protocol}.
The procedure is: we first test the policy trained on the 2.5~kg dumbbell to pick-and-place the 2.5~kg dumbbell (Window 1), and perform this task for multiple times, then we switch to the 5~kg dumbbell without any prior information (Window 2).
The overall visualization of the whole process is shown in \cref{fig:wrench_traj_vis}, where we can see that the internal model successfully learns to predict the external wrench induced by the 2.5~kg dumbbell in Window 1, and keep this internal representation until the load change is detected in Window 2, where the internal model then rapidly re-adapts to the new load in Window 2 upon detecting the load discrepancy.

\begin{figure}[h]
  \centering
  \includegraphics[width=0.8\linewidth]{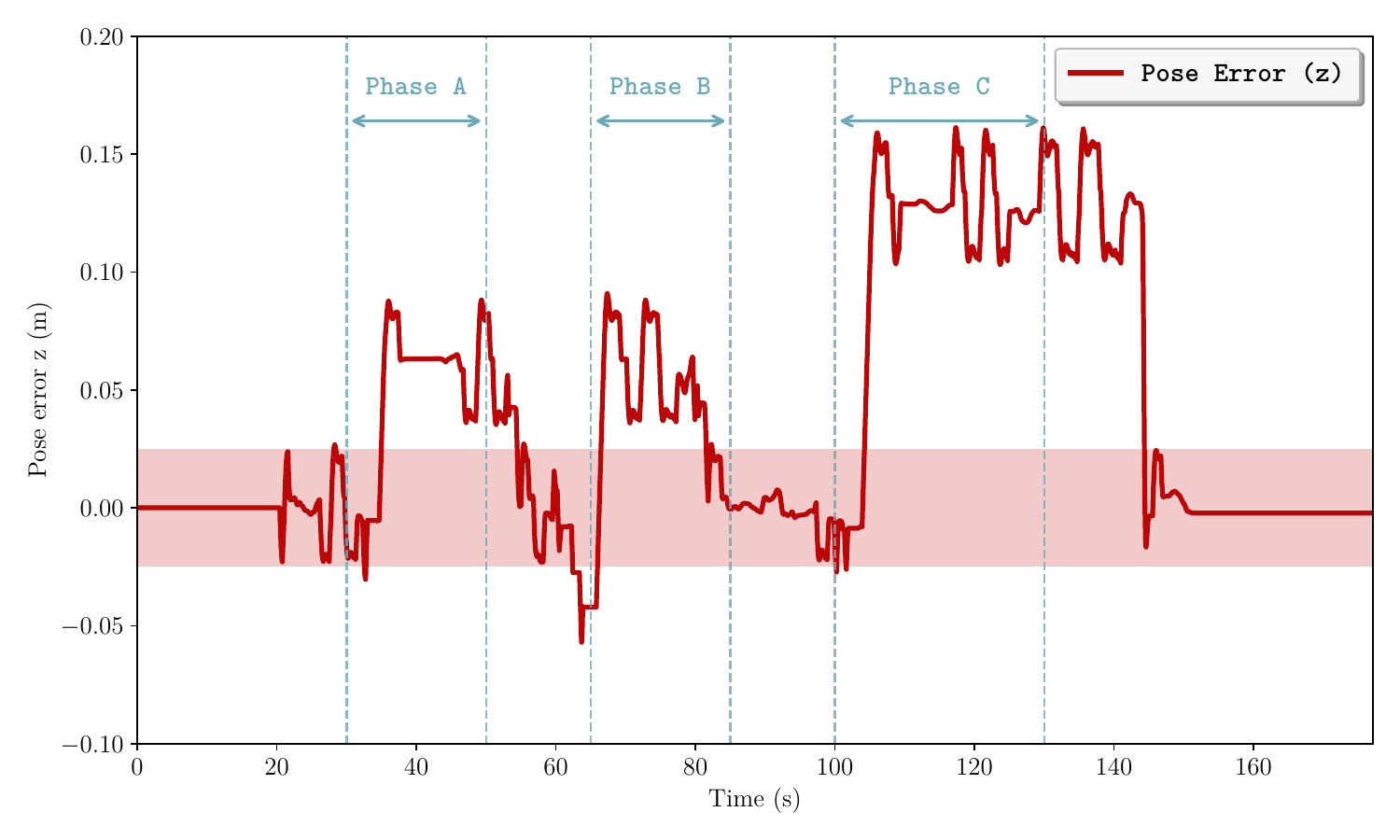}
  \caption{\textbf{Baseline Impedance Control.} Significant steady-state error is observed across all phases due to the lack of load compensation.}
  \label{fig:impedance_pose_err}
\end{figure}

\begin{figure}[h]
  \centering
  \includegraphics[width=0.8\linewidth]{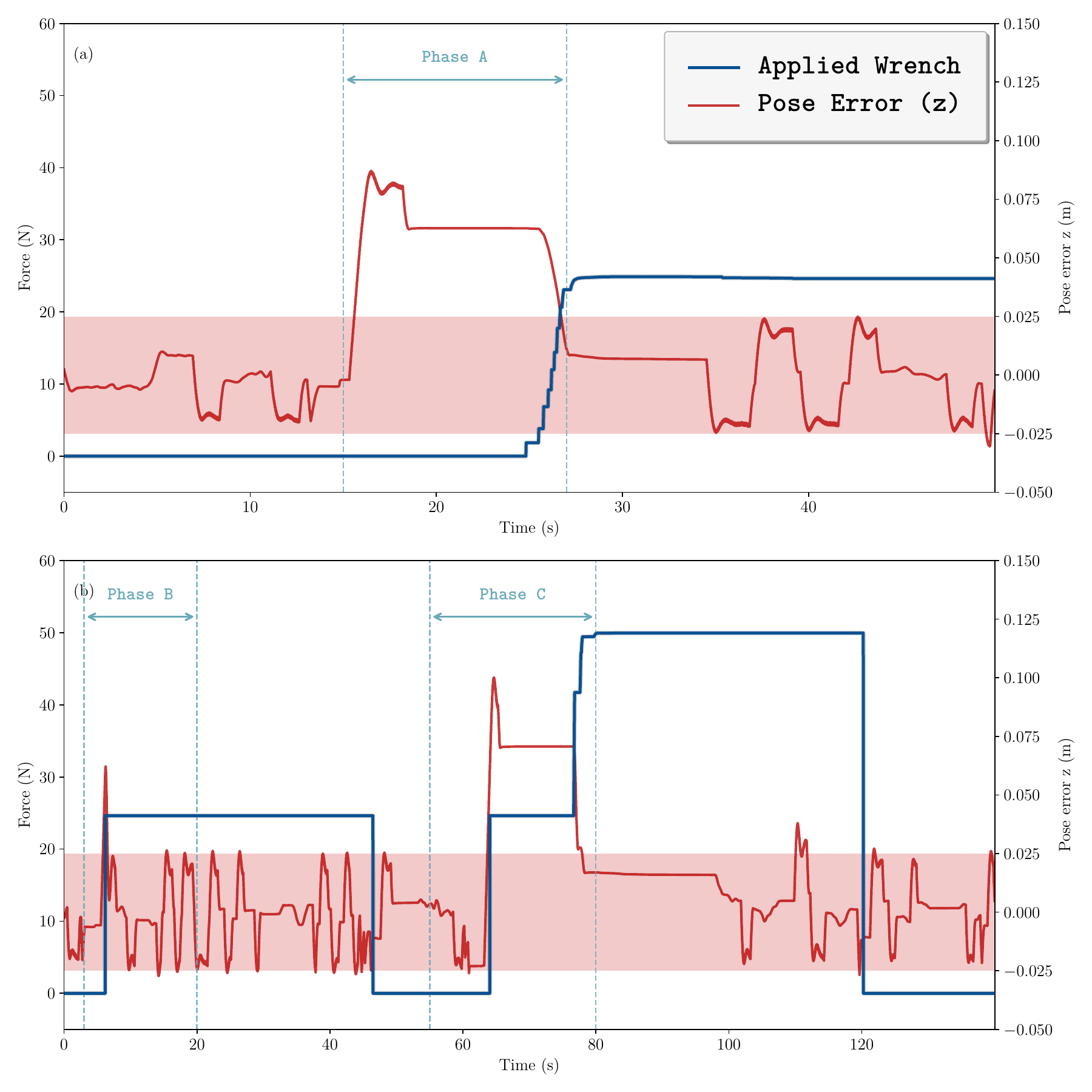}
  \caption{\textbf{\method Performance.} The feedforward wrench effectively cancels external loads, maintaining pose error within the noise floor (red region).}
  \label{fig:pose_err_and_applied_wrench}
\end{figure}

Also, we compare the pose tracking performance of the baseline impedance controller and \method in the real-world experiments.
As shown in \cref{fig:impedance_pose_err}, the baseline impedance controller exhibits significant steady-state pose tracking error across all phases due to the lack of load compensation, while \method effectively cancels the external loads through feedforward compensation, maintaining pose error within the noise floor (red region) as shown in \cref{fig:pose_err_and_applied_wrench}.
This demonstrates the effectiveness of the internal model learning and predictive control mechanism in \method for handling varying and unknown object dynamics in real-world manipulation tasks.

\end{document}